# Otimização de Redes Neurais através de Algoritmos Genéticos Celulares

**Anderson Paulo da Silva, Teresa Bernarda Ludermir**

Centro de Informática – Universidade Federal de Pernambuco - UFPE
{aps3, tbl}@cin.ufpe.br

**Abstract –** This works proposes a methodology to searching for automatically Artificial Neural Networks (ANN) by using Cellular Genetic Algorithm (CGA). The goal of this methodology is to find compact networks whit good performance for classification problems. The main reason for developing this work is centered at the difficulties of configuring compact ANNs with good performance rating. The use of CGAs aims at seeking the components of the RNA in the same way that a common Genetic Algorithm (GA), but it has the differential of incorporating a Cellular Automaton (CA) to give location for the GA individuals. The location imposed by the CA aims to control the spread of solutions in the populations to maintain the genetic diversity for longer time. This genetic diversity is important for obtain good results with the GAs.

**Resumo -** Neste trabalho é proposto um método de busca automática por Redes Neurais Artificiais (RNA) através de Algoritmos Genéticos Celulares (AGC). O objetivo deste método é encontrar redes compactas e com bom desempenho de aprendizado em problemas de classificação. Um dos principais motivos para o desenvolvimento deste método está na difícil tarefa de configurar RNAs compactas com bom desempenho. O uso de AGCs visa buscar pelos componentes das RNAs da mesma forma que um Algoritmo Genético (AG) comum, mas possui o diferencial de incorporar um Autômato Celular (AC) que irá dar localidade aos indivíduos do AG. Essa localidade imposta pelo AC visa controlar a dispersão de soluções dentro da população de forma a manter a diversidade genética por mais tempo. Tal diversidade genética é importante para que bons resultados possam a ser obtidos por AGs.

# 1 Introdução

As redes neurais artificiais (RNAs) têm sido usadas como ferramentas em diversas pesquisas, tais como reconhecimento de padrões [Bishop, 1995], processamento de sinais [Masters, 1994] e predição de séries temporais [Prudencio, 2004]. A sua grande aplicação se dá devido a sua capacidade de induzir uma hipótese, que seja verdadeira para todos (ou muitos) exemplos de uma classe, na qual apenas alguns exemplos são vistos durante um ciclo de treinamento. No entanto, esse sucesso de aplicações com RNAs, depende da correta configuração de seus parâmetros que farão a rede ter o desempenho desejado ou próximo a este. Tais parâmetros dizem respeito à quantidade de camadas escondidas da RNA, à quantidade de nodos em cada camada escondida, aos valores dos pesos associados às conexões entre os nodos, às funções de ativação aplicadas, à taxa de aprendizado da rede, ao algoritmo de aprendizado aplicado e aos parâmetros associados a cada tipo de algoritmo de aprendizado. O problema de encontrar a configuração adequada de parâmetros de RNAs ocorre a cada vez que um novo domínio de aprendizado é posto para a RNA. Cada problema, a ser examinado por uma RNA, acarreta em uma configuração diferente, ou seja, se o domínio do problema mudar, a rede deve ser novamente reconfigurada. Além disso, também existe a necessidade de conhecimento *a priori* sobre o domínio do problema e a presença de um especialista quando não há tais conhecimentos [Almeida and Ludermir, 2007].

Em geral, a configuração de RNAs é feita manualmente, na qual a busca é feita mediante técnicas de tentativa e erro. Assim, são realizadas sucessivas tentativas de ajustes dos parâmetros das RNAs com o intuito de encontrar a configuração que traga resultados satisfatórios. Dessa forma, a busca manual acarreta na necessidade de esforço, experiência e muito tempo de trabalho para encontrar a correta configuração, uma vez que o conjunto de dados a ser ajustado é grande e a quantidade de possibilidades de configurações a ser encontrada manualmente é extensa, o que torna tal tarefa cansativa, pouco produtiva e tendenciosa a erros [Almeida, 2007; Gomes, 2013].

Alguns autores defendem a aplicação de métodos automáticos para a obtenção da correta configuração de RNAs [Bullinaria, 2005]. Entre esses meios, os mais difundidos são os Algoritmos Evolucionários (AEs), que visam "evoluir" as soluções do problema em questão a partir de uma solução inicial (ou um conjunto de soluções iniciais) [Abraham, 2004; Pedrajas, 2005]. AEs aplicam um processo interativo, que trabalha com um conjunto *P* de indivíduos, o que compõe uma população, a qual se aplica uma série de operadores de variação que permitem a evolução e melhoria dos seus resultados. Existem várias técnicas pertencentes ao conjunto dos AEs, como os Algoritmos Genéticos (AGs) que são técnicas com inspiração em processos naturais de evolução (principalmente através de conceitos de biologia evolutiva) [Linden, 2006; Ferreira, 2009], o Enxame de Partículas (do inglês PSO – Particle Swarm Optimization), inspirado no comportamento social de bandos de aves, a Colônia de Formigas (do inglês ACO – Ant Colony Optimization), que é inspirado na observação do comportamento das formigas ao saírem de sua colônia para encontrar comida [Selvi, 2010; Carvalho, 2006; Figueiredo, 2012], além de outras técnicas.

Os AGs são o tipo de aplicação evolucionária mais utilizada com sucesso em conjunto com RNAs. AGs são técnicas evolucionárias baseadas em conceitos de biologia evolutiva como herança, mutação, seleção natural das espécies e outros. A idéia dos AGs é inspirada nas teorias de *Evolução Natural das Espécies,* de *Charles Darwin* [Darwin, 1859].

Quando utilizadas técnicas evolucionárias (entre elas AGs) em conjunto com RNAs, na busca dos componentes da própria RNA, temos as redes neurais artificiais evolucionárias (RNAEs) , que foram definidas por [Yao, 1999]. As RNAEs conseguem ter um melhor aproveitamento (exploitation) e exploração (exploration) de grande número de aspectos e componentes necessários (como pesos, arquiteturas e funções de transição) para a construção de RNAs com bom poder de generalização. Elas utilizam os AGs ao explorar o espaço de soluções para inicializar os seus próprios parâmetros. Assim, por meio dessa união, é possível obter redes neurais com poder de generalização adequado aos problemas propostos, visto que há a aplicação de duas formas de busca em conjunto: a global e a local dentro de espaço de soluções.

No entanto, na construção RNAEs também há restrições. Geralmente esse processo é custoso e possui uma série de peculiaridades de acordo com a técnica evolucionária aplicada. O uso de AGs, por exemplo, possui questões importantes a serem levantadas, como a possibilidade de convergência da busca, dentro de um ponto de ótimo local no espaço de pesquisa. Tal erro pode ocorrer por uma série de motivos, mas, dentre estes, a evolução de indivíduos sem o correto controle, é o mais comum. Quando evoluímos os indivíduos dos AGs sem os devidos cuidados, é possível o desenvolvimento de super indivíduos, que possuirão sempre a mesma carga genética (ou uma carga genética muito parecida). Assim, estes indivíduos irão predominar em relação aos demais, o que traz a convergência genética à população e reduz a sua diversidade. Uma vez que a população de indivíduos não tenha uma boa diversidade, as chances de encontrar soluções diferentes (ou até mesmo a solução ideal) são reduzidas, pois a diversidade genética é de fundamental importância para que o AG encontre boas soluções. Desse modo, ao utilizar AGs, para buscar por configurações ideais de RNAs, também é importante encontrar formas de tornar esse processo mais eficiente e menos propenso a falhas, como a convergência genética. Dessa forma, o presente trabalho baseia-se nos propósitos das RNAEs, nas pesquisas acerca de metodologias evolucionárias para obtenção de RNAs quase-ótimas, no crescente interesse por estas metodologias e nas novas formas de manutenção da população de AGs para obtenção de melhores resultados.

O presente trabalho está organizado da seguinte forma: Seção 2 há uma breve explanação sobre redes neurais e algoritmos genéticos; A Seção 3 explica um pouco sobre os algoritmos evolucionários celulares; A Seção 4 aborda o método proposto; Seção 5 os resultados experimentais e Seção 6 as conclusões do trabalho.

## 2 Redes Neurais Artificiais e Algoritmos Genéticos

As RNAs são definidas como sistemas paralelos distribuídos por unidades de processamento simples (nodos), que calculam determinadas funções matemáticas normalmente não-lineares [Braga et al.2000]. Tais unidades de processamento são dispostas em uma ou mais camadas e interligadas por um grande número de conexões, geralmente unidirecionais. Na maioria dos modelos, estas conexões estão associadas a pesos, os quais armazenam o conhecimento representado no modelo e servem para ponderar a entrada recebida por cada neurônio da rede. As RNAs necessitam passar por um processo de aprendizado, em que um conjunto de exemplo é apresentado à rede. A capacidade de aprender, por meio de exemplos, e de generalizar a informação aprendida é, sem dúvida, o atrativo principal da solução de problemas mediante RNAs [Braga et al. 2000]. Além disso, as RNAs podem ser utilizadas como aproximadores universais de funções multivariáveis e também possuírem a capacidade de auto-organização e de processamento temporal [Haykin, 1999]. Apesar de suas vantagens e de ter uma grande quantidade de aplicações, as RNAs também possuem seus problemas. Um dos principais problemas, no uso de RNAs, consiste em ajustar corretamente um grande número de parâmetros, os quais precisam ser corretamente ajustados para que a aplicação das RNAs possa obter bons resultados e atingir as expectativas desejadas. Tais parâmetros dizem respeito ao número de camadas escondidas, quantidade de neurônios por camada, algoritmos de treinamento empregados, funções de ativação, taxas de aprendizado etc.

Entre as técnicas evolucionárias os AGs são os mais comumente aplicados em conjunto com RNAs, formando as redes neurais evolucionárias (RNAEs) definidas por Yao [Yao, 1999]. As RNAEs possuem capacidade de explorar o espaço de soluções em busca do ponto de ótimo global através do AG e em seguida buscar pelo ponto de ótimo local com o uso da RNA, tornando assim, a busca mais efetiva. Contudo, o uso de RNAs e AGs para otimização de parâmetros se diferencia de metodologia para metodologia. Em alguns casos todos os parâmetros das RNAs podem ser procurados por um AG em outros é possível que o AG busque apenas pelos pesos iniciais da rede. De forma geral, o uso de RNAEs tem trazido bons resultados, seja utilizando o AG para uma pesquisa completa por todos os parâmetros das RNAs [Minku, 2008] ou por apenas parte dos parâmetros. No entanto, mesmo trazendo estes resultados, a aplicação de RNAEs é um processo custoso que possui uma série de peculiaridades que necessitam atenção. A aplicação do AG, por exemplo, possui questões importantes a serem levantadas como a possibilidade de convergência genética da população dentro de um ponto que não é o ótimo global no espaço de busca. Este tipo de erro ocorre mais comumente devido à perda da diversidade causada pelo operador de seleção que constantemente reduz a variedade dos cromossomos, além das perturbações de boas sub-soluções pelas operações de cruzamento e mutação [Linden, 2012; Carvalho, 2003]. Em geral a evolução dos indivíduos sem o correto controle é a causa mais comum para este tipo de falha. Quando evoluímos indivíduos do AG sem o devido controle, é possível o desenvolvimento de super indivíduos que irão se replicar dentro da população, reduzindo a diversidade genética. Uma vez que a população de indivíduos não tenha diversidade, as chances de encontrar soluções diferentes (ou até mesmo a solução ideal) são reduzidas, pois é de fundamental importância que o AG tenha uma boa diversidades dentro de sua população para poder encontrar boas soluções.

Dessa forma, torna-se importante encontrar meios de manter a diversidade populacional dos AGs durante a pesquisa evolucionária, evitando os possíveis erros de convergência genética.

## 3   Algoritmos Evolucionários Celulares

Para contornar os problemas de convergência genética da população, muitos autores propõem modificações nos operadores genéticos dos AGs, como as várias formas de seleção, cruzamento e mutação (como é o caso da seleção por torneio, que tenta tornar o critério de seleção menos elitista que a tradicional seleção por roleta) [Linden, 2006]. Outros autores, por sua vez, têm estruturado a população dos AGs mediante a definição de algum critério de vizinhança entre as soluções manipuladas pelo algoritmo [Nebro, et al. 2006]. Uma das formas encontradas, para amenizar o problema de convergência genética, causado pelos operadores dos AGs, foi proposta por Cao e Wu, no seu trabalho intitulado "A cellular automata based genetic algorithm and its application in machine design optimization" [Cao and Wu, 1998], na qual os pesquisadores introduziram autômatos celulares (ACs) para realizar a localização e vizinhança na estrutura da população dos algoritmos genéticos.

Um AC é um modelo discreto estudado na teoria da computabilidade, matemática e biologia teórica. Consiste de uma grelha finita e regular de células, cada uma podendo estar em um número finito de estados, que variam de acordo com regras determinísticas. A grelha pode ser em qualquer número finito de dimensões. O tempo também é discreto e o estado de cada célula no tempo t é uma função do estado no tempo t − 1de um número finito de células na sua vizinhança. Essa vizinhança corresponde a uma determinada seleção de células próximas (podendo eventualmente incluir a própria célula). Todas as células evoluem segundo a mesma regra de atualização, baseada nos valores de suas células vizinhas. Cada vez que as regras são aplicadas à grelha completa, uma nova geração é produzida. Os autômatos celulares foram introduzidos por Von Neumann e Ulam [Von Neumann, 1966 ] como modelos para estudar processos de crescimento e auto reprodução. Qualquer sistema com muitos elementos idênticos que interagem local e deterministicamente pode ser modelado usando autômatos celulares.

Assim, a saída encontrada em [Cao and Wu, 1998] foi controlar a seleção com base no grid do AC para evitar a perda rápida da diversidade durante a pesquisa genética, baseando-se na distribuição paralela das populações, na qual a ideia de isolar os indivíduos possibilita grande diferenciação genética [Wright, 1943]. Em alguns casos [Alba and Troya, 2002], estes algoritmos utilizando populações descentralizadas conseguem prover uma melhor amostragem do espaço de busca e também melhorar o tempo de execução. Esse tipo de combinação entre AGs e ACs é conhecido como algoritmo genético celular (AGC), que, por sua vez, pertence a um subconjunto dos algoritmos evolucionários chamados de algoritmos evolucionários celulares (AEC).

Um AGC, apesar de utilizar todos os conceitos de um AG comum, possui diferenças que o ajudam a melhorar sua capacidade de busca no espaço de soluções. As principais diferenças entre um AGC e um convencional AG reside em três aspectos fundamentais [Orozco-Monteagudo et al. 2006]:

- Existe um espaço de distribuição que determina o processo de reprodução;
- O processo de seleção é feito para cada elemento da população (apenas no caso de AGCs síncronos, nas quais os elementos, mais aptos ou não, possuem cada um, sua oportunidade de ser selecionado) e os candidatos para cruzamento serão apenas aqueles dentro de uma determinada vizinhança no autômato celular;
- A substituição dos indivíduos da população pode ser feita de duas formas:
  - Por AG geracional, no qual os descendentes sempre substituem seus progenitores;
  - Por elitismo, no qual os descendentes só substituem seus progenitores se eles forem mais aptos.

Tais características garantem uma melhor representação da população no espaço de busca, o que faz o mecanismo de reprodução menos elitista, na qual os indivíduos competem apenas com os seus vizinhos mais próximos.

Nos AECs o conceito de vizinhança é muito utilizado, de modo que um indivíduo interage apenas com seus vizinhos mais próximos. Através dessa restrição de interação, os

AECs conseguem melhorar o comportamento numérico do algoritmo pela estruturação da população, o que consegue, em geral, melhores taxas e tempos de execução. Esse tipo de algoritmo pode formar indivíduos similares em partes distintas da grelha, o que forma os chamados *nichos*. Os nichos funcionam como subpopulações discretas (ilhas). No entanto, não existem fronteiras entre os nichos, por isso é possível a proliferação de tais soluções por toda a população. Assim, os nichos mais próximos podem influenciar mais rapidamente uma determinada região, enquanto os nichos mais distantes transmitem suas soluções mais lentamente [Dorronsoro, 2006].

Em um AEC, a população está normalmente em uma grelha bidimensional, apesar de não haver restrições para a construção de métodos com mais dimensões. Nesta grelha, os indivíduos, que se encontram em uma das bordas, estão conectados aos indivíduos da outra borda, formando uma grelha toroidal. Assim, todos os indivíduos possuem exatamente o mesmo número de vizinhos.

Ao visualizar a estruturação da população de AGs dessa forma, podemos dizer que em um AG comum, sua população estaria como um grafo completamente conectado, onde os vértices são os indivíduos e as arestas as suas relações, enquanto em um AG celular, o uso do AC acaba restringindo o conjunto de relacionamentos entre os indivíduos, isso deixa que as interações aconteçam apenas com os vizinhos mais próximos, como num grafo de estrutura. Dessa forma, as soluções encontradas são lentamente difundidas e manutenção da diversidade acontece mesmo entre os vizinhos mais próximos [Alba and Dorronsoro, 2008].

A figura 1 mostra os dois tipos de grafos citados, à esquerda, há um grafo completamente conectado, representando um AG comum e, à direita um de estrutura, que representa um AGC.

Nos AGCs, cada célula do AC possui um indivíduo do AG e os operadores genéticos (seleção, reprodução e mutação) são aplicados na vizinhança de um determinado cromossomo escolhido no grid. Como dito anteriormente, a evolução natural aqui se faz voltada para o ponto de vista do indivíduo, ou seja, só há cruzamento entre aqueles mais próximos, de forma a simular a influência entre os indivíduos numa sociedade. As interações da população dependem de sua topologia de vizinhança. A vizinhança de um ponto particular da grelha (onde há um indivíduo) se define em termos da distância de Manhattan entre o ponto considerado da malha e os outros pontos da população [Alba and Troya, 2002]. A distância de Manhattan é definida como sendo o cálculo da distância que seria percorrida para ir de um ponto $a$ (de posição $x1, y1$) até um ponto $b$ (de posição $x2, y2$) seguindo um caminho em grade, tal distância formula-se como se fosse de seus componentes correspondentes conforme a fórmula 1.

$$d = \sum_{i=1}^{n} |x_i - y_i| \qquad (1)$$

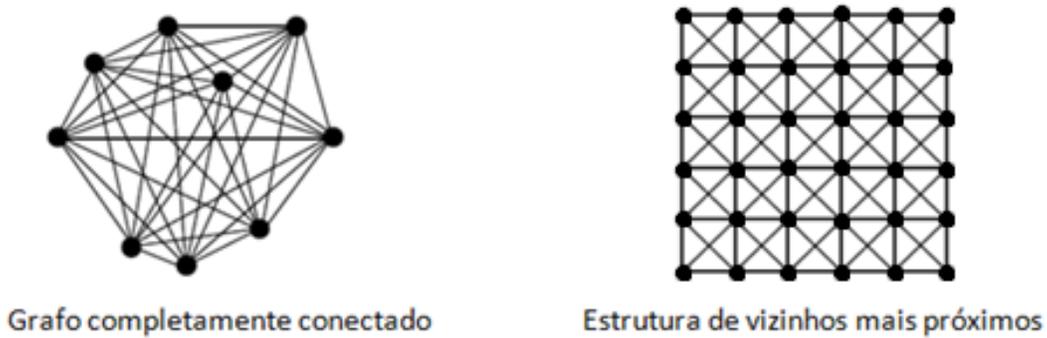

**Figura 1 -** Grafo Completamente Conectado e Grafo de Estrutura

A distância de Manhattan encontra os pontos mais próximos de um dado local por meio de caminhos feitos nos eixos $x$ e $y$ do grid. Dessa forma, é possível obter a vizinhança de um determinado ponto da grelha através dos vértices em que os indivíduos se encontram.

Cada ponto do grid possui uma vizinhança que se sobrepõe com a vizinhança dos indivíduos mais próximos, o que proporciona assim a difusão das soluções. Cada vizinhança do grid possui o mesmo tamanho e a mesma forma. A figura 2 mostra algumas representações de vizinhanças mais facilmente encontradas na literatura [Alba and Dorronsoro, 2008].

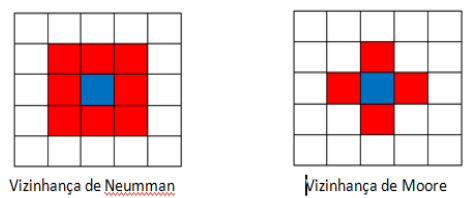

**Figura 2** - Tipos de vizinhanças

Os AGCs só podem interagir com seus vizinhos no ciclo reprodutor quando são aplicados os operadores de variação (cruzamento e mutação). O indivíduo é substituído por um de seus descendentes gerados, seguindo algum critério (se o descendente for uma solução melhor que o pai, por exemplo).

De forma geral, o uso de vizinhanças pequenas é mais comum devido à sua capacidade de manter a diversidade genética por mais tempo. A figura 3 mostra um exemplo de sobreposição de vizinhanças e a dispersão das soluções [Alba and Dorronsoro, 2008].

Apesar de conseguir fazer uma busca eficaz no espaço de soluções, os AGCs que possuem pequena vizinhança são relativamente mais lentos que um AG comum. Isso se dá, principalmente, por causa de sua restrição de vizinhança, na qual a busca é feita de forma mais lenta, o que visa manter mais possibilidades de cruzamento genético dentro da população.

De forma geral, os algoritmos de seleção local são quantitativamente mais fracos que os seus homólogos globais. Isso acontece devido ao fato de que, nos algoritmos de seleção local, é necessário mais tempo para que os bons indivíduos se propaguem por toda população, uma vez que essa propagação só pode ser feita por meio de uma vizinhança restrita.

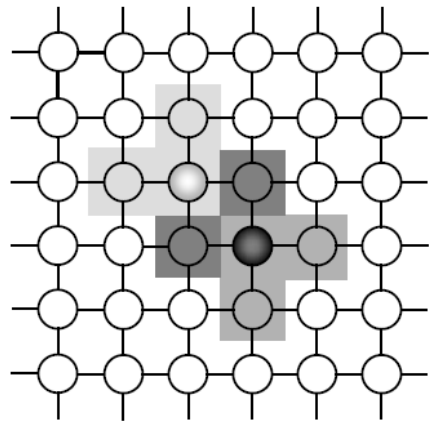

**Figura 3** - Sobreposiçao de vizinhanças e dispersão de soluções

Os AGCs se dividem em dois tipos: Síncronos e Assíncronos. Se o ciclo reprodutor é aplicado a toda população de uma única vez (ao mesmo tempo), chamamos o AGC de síncrono, uma vez que os indivíduos que irão participar da próxima geração são formados todos de uma vez em paralelo. Se o ciclo reprodutor é aplicado sequencialmente seguindo algum determinado critério ao invés de atualizar todos os indivíduos de uma vez, chamamos o AGC de assíncrono [Alba and Dorronsoro, 2008].

Neste trabalho serão utilizados os dois tipos de AGCs, Síncrono e Assíncrono (com ciclo reprodutor feito através de uma escolha uniforme, onde as células são selecionadas com probabilidade uniforme e com reposição (um indivíduo pode ser escolhido mais de uma vez no ciclo reprodutor).

# 4 Algoritmos Genéticos Celulares para otimização de RNAs

Esta seção apresenta o método para otimização de RNAs mediante AGCs, que é baseado, principalmente, na implementação feita em [Almeida, 2007]. A principal diferença desse método é o uso de AGCs que, diferente de um AG comum, trabalham e proporcionam localidade aos indivíduos por meio de um AC, que rege as possibilidades de cruzamento genético. Assim, o AGC contribui com a sua boa capacidade de manutenção da diversidade genética [Silva et al. 2011; Silva, 2021]. Além disso, estes AGCs utilizam os operadores genéticos propostos em [Almeida, 2007], que fazem uso de uma codificação real dos parâmetros das RNAs. O uso desses operadores tem como principal vantagem dispensar a etapa de codificação e decodificação dos cromossomos no processo evolutivo, o que torna a busca evolucionária mais rápida e direta.

O método proposto divide os parâmetros da RNA em três camadas distintas, mas que estão relacionadas entre si, por meio de uma ligação com a camada anterior. Estas camadas dizem respeito a três populações utilizadas pelos AGCs, de forma que cada camada possui um AGC diferente que irá utilizar um AC para definir as regras de cruzamento entre os indivíduos da população. O uso dos AGCs visa buscar um ponto de ótimo global, o qual define redes com melhores configurações iniciais para posterior refinamento da pesquisa por intermédio da busca local feita pela própria RNA mediante algoritmo de aprendizagem.

Esta pesquisa é feita da seguinte forma:

Na camada mais baixa é feita a busca evolucionária por pesos iniciais da RNA (População de Pesos Iniciais – PPI);

Na camada intermediária ocorre a busca evolucionária por arquiteturas e funções de ativação (População de Arquitetura e Funções de Ativação – PAF). Nessa busca estão incluídos o número de camadas escondidas e o número de neurônios por camada;

Na camada superior, é feita a busca evolucionária por regras de aprendizagem, ou seja, os parâmetros dos algoritmos de treinamento (População de Regras de Aprendizagem – PRA).

Assim, apesar de cada AGC trabalhar com seus próprios cromossomos, é possível dizer que eles trabalham com parte de um cromossomo maior, que codifica várias populações diferentes de configurações de RNAS. Cada camada é responsável por otimizar uma parte da rede por meio de um AGC diferente (nos trabalhos anteriormente citados essa busca era feita por intermédio de AGs comuns). A divisão dessas camadas e a forma como elas estão ligadas entre si podem ser observadas na figura 4.

## 4.1 Busca evolucionária por pesos iniciais (BEP)

O método proposto busca por RNAs totalmente conectadas por meio da codificação real dos parâmetros das redes. O uso desse tipo de codificação facilita, principalmente, o trabalho com os pesos das redes, uma vez que a tradicional codificação binária pode gerar cadeias de bits demasiadamente grandes, o que pode proporcionar o indevido controle das cadeias. Além disso, o processo de codificação e decodificação de cadeias binárias é custoso e torna esse tipo de aplicação com baixa viabilidade para os pesos das RNAs.

Assim, cada indivíduo da população de pesos iniciais (PPI) trabalha com matrizes de pesos de uma rede com arquitetura pré-definida. O uso destas matrizes voltadas para uma determinada arquitetura, facilita a manipulação dos pesos, uma vez que todos os indivíduos terão sempre o mesmo tamanho.

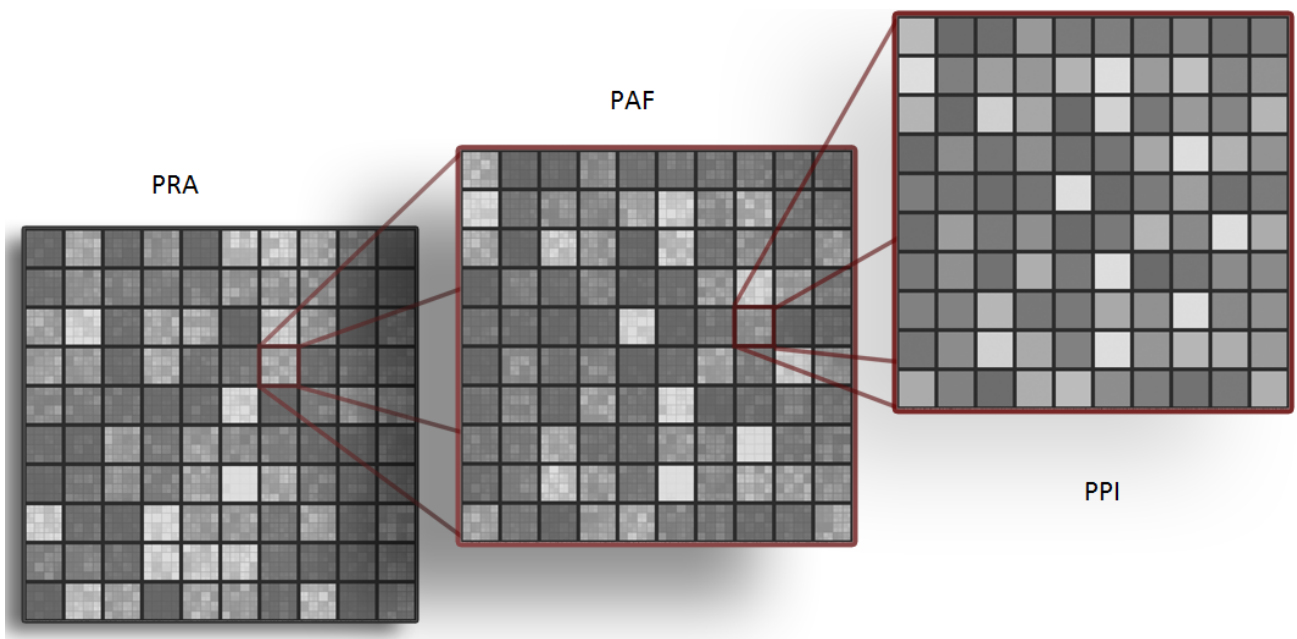

**Figura 4** - Camadas de pesquisa

A busca evolucionária começa com a etapa de seleção dos indivíduos para o cruzamento, para que, em seguida, sejam produzidos os descentes e, por fim, sejam selecionados os sobreviventes de cada geração evolucionária. Uma vez que todas as populações serão inseridas dentro de um AC, que irá aplicar os operadores genéticos de acordo com regras de vizinhança, a seleção dos indivíduos, para cruzamento e seleção de sobreviventes, é feita com a mesma regra, o restará diferenças apenas nos operadores de cruzamento e mutação. Assim, o procedimento de seleção de indivíduos, feito pelo algoritmo 1, é o mesmo utilizado para selecionar todas as populações (PPI, PAF e PRA) para um AGC Síncrono.

A população inicial de indivíduos da PPI é inicialmente gerada, de forma aleatória, a partir de uma distribuição uniforme de valores entre [-0.05, 0.05]. O cálculo de aptidão dos indivíduos da PPI é obtido a partir do erro médio quadrático Mean Squared Erro (MSE) da rede no conjunto de treinamento. A escolha dessa medida de erro foi utilizada com o propósito de reduzir o custo computacional em relação a outras formas de medição.

**Algoritmo 1**: Etapas de evolução dos indivíduos com AGC Síncrono

**Entrada**: **pp**$_k$, // população inicial gerada aleatóriamente de tamanho quadrático *k e já avaliada*

**Saída**: **pp**$_n$

1 **início**
2    Gerar um AC $At$ de dimensões: linhas = $\sqrt{k}$ e colunas = $\sqrt{k}$
3    Inserir em $At$ a população **pp**$_k$
4    Criar outro AC $At_{aux}$ idêntico a $At$
5    **para** *linhas* ← 1 **até** $At.linhas$
6      **para** *colunas* ← 1 **até** $At.colunas$
7        $popAux$ ← selecionaVizinhos($At$, linhas, colunas)
8        $popCruz \leftarrow Cruzamento(popAux, w)$
9        $popMut \leftarrow mutacao(popCruz, n, m, w)$
10       $popMut \leftarrow avalia(popMut)$
11       $At_{aux} \leftarrow replacer(At, popMut, linhas, colunas)$
12    $At \leftarrow At_{aux}$
13    **pp**$_n$ ← $At$
14 **fim**

No algoritmo 1, a população **pp**$_k$, que foi previamente gerada e avaliada, é utilizada como entrada. Nesse momento, é criado um AC $At$, que comporte a população dentro de suas linhas e colunas (justamente por isso o tamanho *k* da população deve ser quadrático). Uma vez inserida a população dentro de $At$, é criado também um AC $At_{aux}$, que irá servir como auxiliar para que o cruzamento de todos os indivíduos aconteçam no mesmo momento da evolução (diferente dos AGCs assíncronos). Em seguida, inicia-se uma caminhada por todas as posições de $At$ (linhas 5 e 6), para fazer a seleção dos vizinhos de cada posição (linha 7). Uma vez selecionados os quatro vizinhos de cada posição (vizinhança de Moore), esses irão para a etapa de cruzamento (linha 8) e, em seguida, para a etapa de mutação, na qual uma quantidade de indivíduos sofrerão mutação de acordo com a taxa *m* de mutação (linha 9). Na linha 10, é aplicada a função *replacer* em $At_{aux}$. Esta função faz uma verificação elitista na posição selecionada. Se o melhor entre todos os elementos do vetor $popMut$ tiver melhor aptidão que o elemento da posição atual de $At$, então $At_{aux}$ passará a guardar esse elemento na mesma posição (substituição do pai pelo filho mais apto). Ao término de todo o processo o AC $At$ passa a ser idêntico a $At_{aux}$ (que possui os indivíduos evoluídos) e dele é retirado os indivíduos evoluídos **pp**$_n$ (seja PPI, PAF ou PRA).

O algoritmo 2 é apresenta a evolução de indivíduos para um AGC Assíncrono. O esquema é basicamente o mesmo adotado pelo ACG Síncrono, com a única diferença que a evolução de indivíduos não acontece ao mesmo tempo com todos os elementos do AC. Assim, os indivíduos evoluem em tempos diferentes na população do AC.

---

**Algoritmo 2**: Etapas de evolução dos indivíduos com AGC Assíncrono

**Entrada**: **pp**$_k$, // população inicial gerada aleatóriamente de tamanho quadrático *k e já avaliada*

**Saída**: **pp**$_n$

1 **início**
2    Gerar um AC *At* de dimensões: linhas = $\sqrt{k}$ e colunas = $\sqrt{k}$
3    Inserir em *At* a população **pp**$_k$
4    Criar outro AC $At_{aux}$ idêntico a *At*
5    **para** *linhas* ← 1 **até** *At.linhas*
6      **para** *colunas* ← 1 **até** *At.colunas*
7        $popAux \leftarrow$ selecionaVizinhos( *At*, linhas, colunas)
8        $popCruz \leftarrow Cruzamento(popAux, w)$
9        $popMut \leftarrow mutacao(popCruz, n, m, w)$
10       $popMut \leftarrow avalia(popMut)$
11       $At \leftarrow replacer\ (At,\ popMut, linhas, colunas)$
12   **pp**$_n \leftarrow At$
13 **fim**

---

Uma vez selecionados os indivíduos, o processo de seleção é encerrado e inicia-se o processo de cruzamento entre os selecionados.

A função do operador de cruzamento é gerar um descendente a partir das características obtidas dos pais selecionados. Uma vez que todos os indivíduos da PPI possuem uma mesma arquitetura previamente definida, o tamanho das matrizes de pesos que forma a PPI, possui a mesma dimensão. Assim, o cruzamento da população PPn anteriormente selecionada, que possui cada um *w* matrizes de pesos de uma rede com dimensões *l x c*, ocorre como descrito no algoritmo 3.

**Algoritmo 3:** Operador de cruzamento para indivíduos da PPI

    **Entrada: pp$_n$, w** // pais, número de matrizes na rede
    **Saída: vetFilhos** // filhos

```
1  inicio
2      numFilhos ← 1
3      enquanto numFilhos ≤ (n/2) faça
4          para matriz ← 1: w faça
5              se rand(1) > probs
6                  filho(matriz) ← fatia2(pp(numFilhos).(matriz),
                                          pp(numFilhos + 1).(matriz))
7              senão
8                  filho(matriz) ← fatia3(pp(numFilhos).(matriz),
                                          pp(numFilhos + 1).(matriz))
9          vetFilhos ← filho
10         numFilhos = numFilhos + 2
11 fim
```

O algoritmo 3 inicia o processo definindo o número de filhos para 1 e então inicia o cruzamento pareado de indivíduos do vetor PPn. Este operador possui uma leve modificação em relação ao operador proposto em [Almeida, 2007], que utilizava uma função fatiar para obter uma das metades de cada indivíduo para posteriormente unir duas metades e gerar o novo filho. O algoritmo 3 utiliza o mesmo conceito, mas dependendo de uma probabilidade *probs,* o filho pode ser gerado não por metade, mas por um conjunto formado a partir de três partes das matrizes. O operador clássico implementado em [Almeida, 2007] utiliza duas matrizes **a** e **b** e obtém as metades $a_d$, $a_e$, $b_d$ e $b_e$. Em seguida é definida a nova matriz filha, por exemplo, $w_{filho}$=[$a_d$; $b_e$] ou $w_{filho} = [a_e; b_d]$. No algoritmo 3 há a possibilidade de, utilizando as mesmas matrizes **a** e **b**, obter as partes $a_d, a_c, a_e, b_d, b_c$ e $b_e$, nas quais $a_c$ e $b_c$ são as partes centrais das matrizes. Assim, um filho pode ser obtido de três partes diferentes, por exemplo, $w_{filho}$=[$a_d$; $b_c$; $a_e$] ou $w_{filho} = [a_e; b_c; b_d]$. Tal modificação foi feita apenas com o intuito de dar mais possibilidades de cruzamentos entre os filhos. Após essa etapa, o processo se repete até que todos os filhos tenham sido gerados com base no valor **n/2**, retornando o vetor **vetFilhos,** que possui todos os filhos gerados.

Uma vez que os filhos foram gerados a próxima etapa é a aplicação do operador de mutação. Esta etapa é necessária para que se mantenha a diversidade genética na população entre as gerações. Os indivíduos ainda não possuem uma aptidão em tal etapa e, por isso, passam por um sorteio aleatório que os selecionará para sofrer mutação. No presente trabalho, a taxa de mutação empregada é de *m=*10%, a qual foi obtida como a que mais contribuiu para bons resultados após sucessivas execuções de teste. Vale ressaltar que, apesar de ser uma taxa alta em relação às taxas encontradas na literatura, é pequena quando comparada ao método proposto em [Almeida, 2007], na qual 40% dos indivíduos passavam pelo processo de mutação. Esta alta taxa de mutação demonstra que o método aplicado tinha uma grande convergência genética e, por isso era preciso aplicar o operador de mutação numa grande parte dos indivíduos. A taxa de mutação, por sua vez, não apenas implica na quantidade de indivíduos que sofreão mutação, como também implica em quantas partes do indivíduo sofrerão mutação. No presente trabalho, a taxa de mutação de 10% implica que 10% dos indivíduos sofrerão mutação em 10% de sua composição. Essa técnica é idêntica à técnica proposta em [Almeida, 2007]. O algoritmo 4 define o operador de mutação aplicado nas PPIs. Inicialmente o algoritmo calcula a quantidade de indivíduos que sofreão mutação. Após isso, o algoritmo entra numa estrutura de repetição, em que a cada interação, um novo indivíduo é mutado de acordo com matrizes esparsas que são geradas para cada um. Estas matrizes possuem o mesmo tamanho dos indíviduos da PPI, com *m%* de seus valores entre $fx = [-0.5, 0.5]$ e os demais valores são zero. Tais matrizes esparsas serão somadas as matrizes de pesos das PPIs. Assim, com um valor *m=*10%, teremos 10% da população sofrendo mutação em 10% de sua composição. Após somar as matrizes, o indivíduo é retirado do vetor de filhos e adicionado ao vetor de filhos com mutação. Essa estrutura de repetição termina quando a quantidade de indivíduos, que sofreram mutação, é alcançada e é retornado um vetor **filhosMut,** que possui todos os filhos (com e sem mutação).

O arcabouço geral da busca evolucionárioa por pesos (BEP) acontece conforme explicado no algoritmo 1.

**Algoritmo 4:** Operador de mutação para indivíduos da PPI

**Entrada: vetFilhos,** *n, m,* w // vetor de filhos, número de filhos, taxa de mutação e quantidade de matrizes de pesos

**Saída:** $filhosMut$

1 **inicio**
2     $quantidadeFilhos \leftarrow ((\frac{m}{100}) * n)$
3     **enquanto** $quantidadeFilhos > 0$ **faça**
4        $filho \leftarrow sortear(\boldsymbol{vetFilhos})$
5        **para matriz** $\leftarrow \mathbf{1} : \mathbf{w}$ **faça**
6           $filho.(matriz) \leftarrow filho.(matriz) + gerarMatrizEsparsa(fx, m)$
7        $\boldsymbol{vetFilhos} \leftarrow \boldsymbol{vetFilhos} - filho$
8        $\boldsymbol{filhosMut} \leftarrow filho$
9        $quantidadeFilhos \leftarrow quantidadeFilhos - 1$
10     $\boldsymbol{filhosMut} \leftarrow \boldsymbol{filhosMut} + vetFilhos$
11 **fim**

## 4.2 Busca evolucionária por arquiteturas e funções de ativação (BEAF)

A seleção de indivíduos da PAF é feita de acordo com os critérios dos AGCs do algoritmo 1 ou 2 (síncrono ou assíncrono). Desse modo, a busca realizada por arquiteturas também segue o mesmo critério da BEP, na qual as novas arquiteturas são geradas seguindo características genéticas de seus pais, tais como informações da quantidade de camadas e unidades escondidas ou, até mesmo, as funções de ativação. A BEAF busca as camadas escondidas (uma vez que as camadas de entrada e saída são definidas de acordo com o problema a ser pesquisado), quantidade de neurônios por camada e as funções de ativação. A heterogeneidade das funções de ativação, aplicadas às camadas das RNAs, podem ser diferentes umas das outras, ou seja, uma camada pode ter uma função tangente hiperbólica sigmóide e outra camada pode ter uma função linear, por exemplo. Essa aplicação foi feita devido à ocorrência de estudos nesse sentido, como também pela capacidade de exploração dos benefícios da união entre RNAs lineares e não-lineares, já que uma rede tida como não-linear, que utiliza funções de ativação lineares, é vista como linear [Almeida, 2007].

       Eis as funções de ativação utilizadas durante a pesquisa evolucionária: função linear (*Pure line transfer function*) FL, tangente hiperbólica sigmóide (*Hyperbolic tangent sigmoid transfer function*) FTH e a logarítmica sigmóide (*Logarithmic sigmoid transfer*

*function*) FLS. A linear é utilizada também como função de ativação da camada de saída para todas as redes. A quantidade de camadas escondidas, em cada rede, pode variar de [1,3]. A quantidade de nodos, em cada camada, pode variar de [1,12]. Este número foi escolhido devido à existência de alguns problemas que necessitam de mais neurônios para que a rede consiga uma boa generalização.

Diferente da função de aptidão adotada em [Almeida, 2007], a de aptidão, adotada em todas as populações, foi feita através do cálculo do MSE da rede. Esta função foi escolhida unicamente com o intuito de reduzir o custo gerado pela função proposta que possuía grande número de componentes, além de valores empíricos utilizados.

Dessa maneira, o operador de cruzamento utilizado para as PAFs é executado logo após o AGC selecionar os indivíduos de acordo com o critério de vizinhança adotado. No entanto, após essa seleção, é feita uma busca interna dentro do AC para encontrar PAFs, as quais são estruturalmente iguais. Caso existam duas arquiteturas, que tenham a mesma estrutura (número de camadas e nodos por camada), é realizada a aplicação de uma função, que vai mutar uma dessas arquiteturas. A aplicação desta mutação é feita visando à manutenção da diversidade genética na PAF. O cruzamento das PAFs é realizado de acordo com o algoritmo 5.

O algoritmo de cruzamento da PAF recebe, como parâmetros, a população previamente selecionada pelo AGC e começa o seu processo evolucionário pela definição da quantidade de camadas escondidas que o filho irá ter (linhas 4 a 8). Primeiramente, é definido o número total de camadas a partir do somatório da quantidade de camadas dos respectivos pais. Em seguida, dependendo de uma probabilidade *probs*, o filho irá ter a média de camadas escondidas dos pais arredondadas para cima ou para baixo [Almeida, 2007].

Em seguida, é feita a união das dimensões de cada camada relativa aos pais e de suas funções de transferências. A partir dessa união, será sorteada a quantidade de neurônios por cada camada dos filhos e suas funções de ativação. Esse processo se repete até que a quantidade especificada de cruzamentos seja realizada.

**Algoritmo 5: Operador de cruzamento para indivíduos da PAF**

   Entrada: $paf_n$, $n$ // pais, quantidade de cruzamentos
   Saída:     vetFilhos // filhos

1 **inicio**
2    $numfilhos \leftarrow 1$
3    **enquanto** $numfilhos \leq (n/2)$ **faça**
4       **totalCamadas** $\leftarrow$
         $\mathbf{paf}(numfilhos).\mathbf{numCamEsc} + \mathbf{paf}(numfilhos + 1).\mathbf{numCamEsc}$
5       **se** $\mathbf{rand}(1) > probs$ **faça**
6             $\mathbf{filho}.\mathbf{numCamEsc} \leftarrow \mathbf{ceil}(totalCamadas/2)$
7       **senão**
8          $\mathbf{filho}.\mathbf{numCamEsc} \leftarrow \mathbf{floor}(totalCamadas/2)$
9       $\mathbf{dimensoes} \leftarrow [\mathbf{paf}(numfilhos).\mathbf{dim}; \mathbf{paf}(numfilhos + 1).\mathbf{dim}]$
10      $\mathbf{funcoes} \leftarrow [\mathbf{paf}(numfilhos).\mathbf{func}; \mathbf{paf}(numfilhos + 1).\mathbf{func}]$
11      **para** $elementos \leftarrow 1: \mathbf{dimensoes}_{tamanho}$ **faça**
12         $[\mathbf{filho}.\mathbf{dim}, \mathbf{filho}.\mathbf{func}] \leftarrow \mathbf{sorteia}(\mathbf{dimensoes}, \mathbf{funcoes})$
13      $\mathbf{vetFilhos} \leftarrow \mathbf{filho}$
14      $numfilhos \leftarrow numfilhos + 2$
15 **fim**

Em função da simplicidade do operador de cruzamento adotado, há a possibilidade de serem produzidos indivíduos similares aos seus pais, o que dificulta a busca devido à redução da diversidade genética. Por isso, após o cruzamento é preciso aplicar o operador de mutação para manter a diversidade da população. A mutação de PAFs é feita pelo algoritmo 6 que começa com a definição dos filhos que sofrerão mutação. Em seguida é gerado um número aleatório no intervalo [0,1]. Este número será utilizado como uma probabilidade que indicará se uma arquitetura irá sofrer um incremento ou decremento na quantidade de camadas escondidas ou no número de neurônios escondidos por camada. Na etapa de mutação, são avaliados todos os casos possíveis de quantidade de camadas (1, 2 ou 3). De acordo com uma determinada probabilidade, o indivíduo pode ter sua quantidade de camadas aumentada ou sofrer uma redução de uma ou duas camadas. Caso o indivíduo tenha 1 ou 2 camadas ocultas, há maiores chances de este passar a ter 3 camadas (visando reduzir o número de nodos por camada). A probabilidade de redução de camadas só é maior para indivíduos com 3 camadas ocultas (tendo esse maiores chances de passar a ter 2 camadas ocultas). De forma geral, essa etapa prioriza gerar indivíduos com mais camadas e menos nodos por camada. Caso a probabilidade não atinja um valor esperado para adicionar ou remover camadas dentro das arquiteturas, ocorrerá uma mudança nas dimensões das camadas, ou seja, a arquitetura será mantida, mas a quantidade de nodos, em cada camada, será modificada através de valores gerados entre $arqval = [-2, 5]$ que serão adicionados às dimensões atuais da arquitetura (tendo no mínimo 1 e no máximo 12 neurônios por camada).

## 4.3 Busca evolucionária por regras de aprendizagem (BERA)

Um algoritmo de aprendizagem para RNAs pode ter diferentes performances quando aplicada a diferentes arquiteturas. O desenvolvimento de algoritmos de aprendizagem, mais fundamentalmente da regra de aprendizagem usada para ajustar os pesos das conexões, depende do tipo de arquitetura trabalhada. Diferentes variações da regra de aprendizagem Hebbiana foram propostas para lidar com diferentes arquiteturas. Porém, o desenvolvimento de uma regra de aprendizagem ótima se torna muito difícil quando se tem pouco conhecimento *a priori* sobre a arquitetura da rede neural, o que acontece com frequência na prática [Yao, 1999]. Sendo assim, o ajuste adaptativo dos parâmetros do BP (como taxa de aprendizagem e termo de momento) por meio da evolução, pode ser considerado como uma primeira tentativa de evolução da regra de aprendizagem, tendo em vista as dificuldades de criação e ajuste de toda uma regra de aprendizagem para cada arquitetura localizada na busca evolucionária [Yao, 1999]. Assim, a busca evolucionária por regras de aprendizagem é feita mediante busca dos parâmetros dos algoritmos de aprendizados aplicados as RNAs. Em cada PRA, há os parâmetros necessários a cada algoritmo de aprendizado aplicado e, dependendo do algoritmo em questão, tais valores irão passar pela busca evolutiva de forma a procurar os parâmetros que irão proporcionar a melhor aptidão para as RNAs.

Dessa forma, a aptidão das PRAs é baseada na aptidão das PAFs que tem como função utilizada, a medida do MSE da rede no conjunto de treinamento. Inicialmente foram utilizados quatro algoritmos de aprendizado para essa etapa da busca evolucionária, mas após vários experimentos ficou perceptível que dois desses algoritmos conseguiam os melhores resultados na maioria dos experimentos (em termos de MSE e sensibilidade a mudanças de arquiteturas). Com isso, visando reduzir o custo computacional do método aplicado, apenas os dois melhores algoritmos de aprendizado foram utilizados: o algoritmo de Gradiente Conjugado Escalonado (SCG) e o Backpropagation (BP).

---

**Algoritmo 6: Operador de mutação para indivíduos da PAF**

**Entrada:** vetFilhos, n, m // vetor com os filhos, número de filhos, taxa de mutação
Saída: **filhosMut** // Filhos mutados

```
1  inicio
2      quantidadeFilhos ← ((m/100) * n) faça
3          enquanto quantidadeFilhos > 0 faça
4              probabilidade ← rand(1)
5              filho ← sortear(vetFilhos)
6                  se filho.numCamEsc = 1 faça
7                      se probabilidade ≥ 0.6 faça
8                          filho ← adicionaCamadas(3)
9                      senão se probabilidade ≥ 0.5 então
10                         filho ← adicionaCamadas(2)
11                     senão
12                         filho ← filho + sorteia(arqval)
13
14                 senão se filho.numCamEsc = 2 então
15                     se probabilidade ≥ 0.6 então
16                         filho ← adicionaCamadas(3)
17                     senão se probabilidade ≥ 0.5 então
18                         filho ← reduzCamadaPara(1)
19                     senão
20                         filho ← filho + sorteia(arqval)
21
22                 senão
23                     se probabilidade ≥ 0.6 então
24                         filho ← reduzCamadasPara(2)
25                     senão se probabilidade ≥ 0.5 então
26                         filho ← reduzCamadasPara(1)
27                     senão
28                         filho ← filho + sorteia(arqval)

29             vetFilhos ← vetFilhos − filho
30             filhosMut ← filho
31             quantidadeFilhos ← quantidadeFilhos − 1
32     filhosMut ← FilhosMut + vetFilhos
10 fim
```

Como os indivíduos da PRA possuem as informações dos parâmetros do SCG e do BP, o cruzamento ocorre pela variação dos valores dos parâmetros dentro da faixa de valores pertencentes aos seus pais. A mutação por sua vez, é executada de acordo com uma adição ou subtração de um percentual dos valores dos filhos. Este percentual de mutação *m* é o mesmo aplicado as demais buscas evolucionárias, ou seja, 10%. Dessa forma, os indivíduos da PRA que passarem pelo processo de mutação terão, dependendo de uma determinada probabilidade, uma adição ou subtração de 10% de seus próprios valores.

### 4.4 Descrição global do método proposto

A busca automática por RNAs utiliza três AGCs responsáveis por otimizar partes distintas das RNAs. No entanto, os AGCs estão interligados entre si, trocando informações das camadas de busca com o intuito de encontrar uma configuração de parâmetros, que gere RNAs compactas com bons desempenhos. Como já mencionando anteriormente, os indivíduos da PRA possuem uma PAF, que, por sua vez, também possui uma PPI. Por meio dessas relações, as trocas de informações são feitas e os parâmetros necessários são encontrados.

O processo geral de busca por RNAs é feito para cada algoritmo de aprendizagem, na qual inicialmente as populações são geradas de forma aleatória, os indivíduos são avaliados e os ciclos de busca evolucionária se iniciam. Dessa forma, numa camada mais interna, é feita a busca por pesos iniciais de acordo com as gerações da BEP.

Em seguida, a busca por PAFs realizada, da mesma forma que a busca por PPIs, de acordo com as gerações de BEAFA e, por último, o processo se repete da mesma forma para a busca por PRAs de acordo com as gerações da BERA.

Esse processo se repete para cada um dos algoritmos de aprendizado aplicado e, ao término dos ciclos de busca, as melhores RNAs são selecionadas.

## 5 Resultados Experimentais
### 5.1 Introdução

Uma vez que o método proposto tenha suas peculiaridades esclarecidas e devidamente apresentadas, é preciso testá-lo para avaliar a sua eficiência experimentalmente. Assim, nessa seção, serão apresentados os meios experimentais voltados para testar o presente trabalho e apresentar os resultados obtidos após todas as execuções necessárias dos algoritmos. Esses resultados são comparados com o método NNGA-DCOD proposto em [Almeida, 2007] e que utiliza os mesmos domínios de problemas usados para testar a abordagem celular proposta neste trabalho, além de uma breve comparação com a busca manual. Cada um dos testes foi executado seguindo um número determinado de gerações para cada AGC proposto e os parâmetros necessários para execução dos testes. A tabela 5.1 apresenta os detalhes dos parâmetros de cada componente da pesquisa utilizado na execução dos experimentos. A seção 5.2 apresenta as bases de dados utilizadas durante os experimentos, juntamente com suas características (atributos e classes); a seção 5.3 explica a forma de execução dos experimentos; a seção 5.4 mostra os resultados obtidos mediante método proposto comparado com o método NNGA-DCOD por meio de execuções em que os parâmetros dos dois métodos são iguais (algoritmos utilizados, gerações de busca evolucionária, método de obtenção dos resultados e o teste estatístico aplicado).

| **Algoritmo 7: Visão geral da busca evolucionária por RNAs** |
|---|
| **Entrada:** Parâmetros a serem otimizados // |
| **Saída: Um conjunto de redes quase − ótimas//** |
| 1 **inicio** |
| 2     Gere aleatoriamente as populações de: Pesos iniciais (PPI), Arquiteturas e funções (PAF) e Regras de aprendizagem (PRA) |
| 3     **para cada algoritmo de aprendizagem faça** |
| 4        **Calcule a aptidão para todos os indivíduos** |
| 5           **para cada geração ∈ BERA faça** |
| 6             **para cada PAF ∈ PRA e geração ∈ BEAFA faça** |
| 7                **para cada PPI ∈ PAF e geração ∈ BEP faça** |
| 8                   Utilize o algoritmo 1 (ou algoritmo 2 dependendo da aplicação sincrona ou assincrona) com os indivíduos da BEP |
| 9             Repita os passos da BEP com os indivíduos da BEAFA |
| 10           Repita os passos da BEP com os indivíduos da BERA |
| 11     Selecionar redes quase-ótimas obtidas com cada algoritmo de aprendizagem |
| 12 **fim** |

## 5.2 As bases experimentais

Os experimentos foram realizados visando comparar os resultados do método celular em relação ao método proposto em [Almeida, 2007] nomeado de NNGA-DCOD. Foram utilizadas dez bases conhecidas do repositório UCI [Newman et al. 1998], são elas: Crédito Alemão, Ionosfera, Veículos, Cavalos, Câncer, Diabetes, Vidros, Coração, Crédito Australiano e Sonar. A descrição de um domínio de problema (base) é composta por um conjunto de parâmetros, que delimitam as fronteiras e mapeiam, por meio de escalas, as informações do problema. Às vezes, as escalas não contribuem para uma boa compreensão da descrição sistemática do problema. Dentre os parâmetros de um domínio de problema, que são chamados de atributos (AT), existe um atributo denominado de Classe (CA), que determina a classe, à qual um dado conjunto de atributos pertence [Almeida, 2007].

A base de dados é composta pelo conjunto de atributos anteriormente mencionado e possui exemplos que abordam cada um dos atributos, incluindo o atributo de classe. Estes exemplos possuem os dados característicos nos atributos que indicam quando um exemplo pertence a uma determinada classe ou não. A base é vista como uma instância do domínio do problema; é a partir dela que a rede tentará aprender e construir uma descrição de todo problema [Almeida, 2007].

Para utilizar tais bases com RNAs, as mesmas precisam ser divididas em três subconjuntos: O conjunto de treinamento (TR), o conjunto de validação (VL) e o conjunto de teste (TE). A rede é treinada de acordo com o conjunto de treinamento. É nessa fase que os pesos entre as conexões da rede são ajustados ao buscar o aprendizado do domínio do problema. O treinamento da rede é interrompido de acordo com o erro do conjunto de validação. Para finalizar, a rede tem sua capacidade de aprendizado testada por intermédio do conjunto de teste. A tabela 5.2 mostra a divisão das bases e explica, de forma sucinta, cada uma das bases. Os domínios escolhidos possuem diferentes quantidades de exemplos: atributos e classes. Essa variedade de características foi escolhida de forma a testar o método proposto em diferentes domínios, tornando possível verificar como o método consegue lidar com diferentes características nos domínios utilizados.

|  | *Parâmetros para* | *Valores* |
|---|---|---|
| **AGCs** | *Modo de codificação* | Real |
|  | *Tipo de seleção* | Elitista |
|  | *Taxa de Mutação* | 10% |
|  | *Número de indivíduos por população (PRA, PAF e PPI)* | 16, 25, 25 |
|  | *Quantidade de gerações (BERA, BEAFA, BEP)* | 16, 5, 3 |
|  | *Cálculo de aptidão* | Baseado no MSE do conj. de treinamento |
| **RNAs** | *Tipo da Rede* | MLP feedforward |
|  | *Algoritmos de aprendizagem empregados* | BP e SCG |
|  | *funções de transferência* | FL (linear), FLS (logarítmica Sigmoid) e FTH (tangente hiperbólica) |
|  | *Número de camada escondidas* | até 3 |
|  | *Número de unidades escondidas* | até 12 |
|  | *Número de épocas de treinamento* | até 50 |
|  | *Intervalo para os pesos iniciais* | [-0.05, 0.05] |
|  | *Função de transferência da última camada* | FL |
| **Algoritmos de aprendizagem** | *BP* |  |
|  | *Taxa de aprendizado* | [0.05, 0.25] |
|  | *Momento* | [0.05, 0.25] |
|  | *SCG* |  |
|  | *Fator de controle para cálculo de aproximação da informação de segunda-ordem* | [0, 0.0001] |
|  | *Fator regulador da falta de definição da matriz hessiana* | [0, 1.0e-06] |

**Tabela 5.1 -** Parâmetros utilizados na pesquisa

A divisão dos conjuntos de dados seguiu o método utilizado em [Cantú-Paz and Kamath, 2005], no qual, a cada iteração os dados, foram aleatoriamente divididos em metades. Uma das metades foi utilizada como entrada para os algoritmos com 70% de seus dados para treinamento e 30% para o conjunto de validação. A outra metade da base foi utilizada como conjunto de teste.

## 5.3 Execução dos Experimentos

De forma geral, os experimentos encontrados na literatura são executados ao utilizar os erros do conjunto de treinamento, validação e teste, além de utilizar um bom número de domínios de problemas [Prechelt, 1994]. Uma vez que os resultados são obtidos, a significância estatística, em relação a outro experimento, deve ser testada. Assim, na comparação de classificação de RNAs, os experimentos são executados ao adotar validação cruzada [Stone, 1978] com *k*-partições do conjunto de dados para estimar a eficiência das redes, além da execução do *t*-teste [Witt and McGrain, 1985] para confirmar se há significância estatística entre os resultados.

De acordo algumas pesquisas, houve a constatação que o uso da validação cruzada com o teste estatístico *t*, aumentou a ocorrência do erro tipo-1: Os resultados são erroneamente classificados como significativamente diferentes mais vezes que o esperado [Dietterich, 1998. Cantú-Paz and Kamath, 2005]. Devido à ocorrência do tipo-1 de erro, neste trabalho foi adotado o mesmo tipo de estratégia utilizada em [Almeida, 2007] que, por sua vez, utilizou a mesma estratégia adotada Dietterich [Dietterich, 1998] e Alpaydin [Alpaydin, 1999], na qual o algoritmo 7 foi executado dez vezes com cinco divisões diferentes do conjunto de dados, usando validação cruzada (5 x 2 *vc*). Para cada duas execuções completas do método proposto, o conjunto de dados é particionado aleatoriamente em metades $D_1$ e $D_2$, onde, na primeira dessas execuções, as redes são treinadas e validadas com $D_1$ e testadas com $D_2$. Na execução seguinte, ocorre o contrário, as redes são treinadas e validadas com $D_2$ e testadas com $D_1$. Para verificar estatisticamente a diferença, é utilizada uma combinação dos valores obtidos nos conjuntos de teste com o teste estatístico *f-teste*. A execução deste teste foi realizada com o auxílio do software BioStat 5.0.

O F é um teste de hipóteses, que é aproximadamente distribuído com cerca de cinco a dez graus de liberdade e rejeita a hipótese nula de que os dois algoritmos têm a mesma taxa de erro com uma significância de 0,05 se F > 4,74. Essa metodologia foi utilizada devido ao fato do método usual (teste T de Student) gerar um aumento de erros do tipo-1, no qual os resultados são erroneamente considerados diferentes com mais freqüência do que o esperado, dado o nível de confiança utilizado. O teste foi aplicado sobre a distribuição dos dez melhores resultados de cada base no conjunto de erros de testes das RNAs.

| Nome do domínio | Composição | | | | | Descrição |
|---|---|---|---|---|---|---|
| | TR | VA | TE | CA | AT | |
| Crédito Alemão | 350 | 150 | 500 | 2 | 24 | Base com dados de possíveis clientes de crédito alemão, classificando como bom ou mal pagador. |
| Ionosfera | 122 | 53 | 176 | 2 | 34 | Dados relativos a elétrons livres na ionosfera. Classifica se existe alguma estrutura de elétrons ou não |
| Veículos | 296 | 127 | 423 | 4 | 18 | Base com dados de silhuetas de quatro tipos de veículos diferentes. |
| Cavalos | 106 | 45 | 182 | 3 | 58 | Dados relativos a cavalos que estão com cólica. Estes dados são utilizados para predizer se os cavalos vão sobreviver, morrer ou se terão que ser sacrificados. |
| Câncer | 244 | 105 | 350 | 2 | 9 | Trata da verificação do tipo de câncer de mama, com a existência de duas classes, se é benigno (65.5% dos casos) ou maligno (34.5%) |
| Diabetes | 269 | 115 | 384 | 2 | 8 | Contém informações de indivíduos de uma tribo indígena. Estas informações são utilizadas para predizer quando apresentam sinais de diabetes. |
| Vidros | 75 | 32 | 107 | 6 | 9 | Possui informações para classificação de 6 tipos de vidros diferentes |
| Coração | 322 | 138 | 460 | 2 | 35 | Trata de verificar do tipo de câncer de mama, com a existência de duas classes, se é benigno (65.5% dos casos) ou maligno (34.5%) |
| Crédito Australiano | 242 | 103 | 345 | 2 | 14 | Base com dados de possíveis clientes de crédito australiano, classificando como bom ou mau pagador. |
| Sonar | 73 | 31 | 104 | 2 | 60 | Base com dados sobre possíveis minas submersas. Classifica entre Minas ou rochas. |

**Tabela 5.2 -** Descrição das bases

Legenda; CA – Classes, AT – Atributos, TR – Tamanho do conjunto de treinamento, VA – Tamanho do conjunto de validação, TE – Tamanho do conjunto de teste.

Assim, seja a hipótese nula H0 e a hipótese alternativa H1:

H0: Os métodos possuem médias iguais

H1: Os métodos possuem médias diferentes

O teste é realizado para verificar se houve negação da hipótese H0 ou não.

## 5.4 Resultados

A tabela 5.3 apresenta os valores médios dos dez melhores resultados obtidos com os métodos celulares (síncrono e assíncrono) e a média dos dez melhores resultados do NNGA-DCOD, juntamente com os seus respectivos desvios para cada algoritmo de aprendizado. Vale ressaltar que, para cada algoritmo de aprendizado, existem dez estimativas de erro de teste e que a média apresentada, corresponde à média destas dez estimativas.

A tabela 5.4 mostra os resultados dos testes estatísticos comparados entre as técnicas celulares e o NNGA-DCOD de acordo com cada algoritmo aplicado em todos os domínios de problemas utilizados. O teste foi feito de forma a avaliar os desempenhos de cada AGC em relação ao NNGA-DCOD utilizando o mesmo tipo de algoritmo. Os símbolos **>** e **<** significam que a comparação obteve valores maiores que 4.74 e menores que 4.74 respectivamente. O símbolo **–** é utilizando apenas para separar os dados, uma vez que não foi feita comparação entre algoritmos diferentes.

De acordo com o teste F os resultados do AGC Síncrono foram superiores ao NNGA-DCOD em todas as aplicações, mostrando que a ferramenta obteve o êxito proposto e conseguindo atingir os resultados desejados. O AGC assíncrono só conseguiu ser completamente superior com o uso do algoritmo SCG, uma vez que ele não conseguiu atingir os resultados desejados em 4 das cinco bases com o algoritmo BP. Contudo, vale ressaltar que suas médias ficaram no pior caso muito próximas ao método NNGA-DCOD.

| VALORES DO MSE (EM %) PARA CADA BASE | | | | | | | |
|---|---|---|---|---|---|---|---|
| Algoritmos / Métodos | | | NNGA-DCOD | | AGC-1 ASSINCRONO | | AGC-2 SÍNCRONO | |
| | | | BP | SCG | BP | SCG | BP | SCG |
| BASES | Crédito Alemão | Média | 24.39 | 20,35 | 22,41 | 16,89 | 22,51 | 16,92 |
| | | Desvio | 0,1425 | 2,3657 | 0,4674 | 0,2221 | 0,3834 | 0,8666 |
| | Ionosfera | Média | 24,31 | 16,63 | 23,63 | 7,66 | 23,81 | 8,26 |
| | | Desvio | 0,4563 | 3,3796 | 0,9057 | 0,9364 | 0,3796 | 0,6057 |
| | Veículos | Média | 24,93 | 21,87 | 24,94 | 9,89 | 24,93 | 9,78 |
| | | Desvio | 0,0278 | 1,391 | 0,0248 | 0,3673 | 0,0164 | 0,3546 |
| | Cavalos | Média | 24,14 | 16.46 | 23,11 | 13,60 | 23,06 | 15.79 |
| | | Desvio | 0,1244 | 0,7961 | 0,2641 | 0,1005 | 0,6402 | 0,4176 |
| | Diabetes | Média | 24,45 | 21,06 | 24,19 | 20,12 | 23,55 | 16,39 |
| | | Desvio | 0,2015 | 1,8118 | 0,1613 | 1,3044 | 0,3898 | 0,2743 |
| | Câncer | Média | 23,98 | 4,12 | 18,44 | 2,09 | 10,96 | 2,01 |
| | | Desvio | 0,5332 | 0,812 | 0,42923 | 0,475 | 0,0214 | 0,34 |
| | Vidros | Média | 13,84 | 11,76 | 13,55 | 11,0052 | 13,12 | 11,02 |
| | | Desvio | 0,1646 | 0,9312 | 0,4548 | 0,2104 | 0,2477 | 0,1107 |
| | Coração | Média | 24,85 | 16,87 | 24,21 | 13,02 | 23,76 | 12,36 |
| | | Desvio | 0,06 | 1,62 | 0,301 | 0,912 | 0,83 | 0,43 |
| | Sonar | Média | 24,96 | 18,07 | 24,95 | 17,49 | 24,93 | 17,66 |
| | | Desvio | 0,05 | 1,365 | 0,095 | 0,2307 | 0,092 | 0,44 |
| | C. Australiano | Média | 24,87 | 11,18 | 24,66 | 10,79 | 24,64 | 10,39 |
| | | Desvio | 0,0766 | 0,8731 | 0,0977 | 0,2937 | 0,09 | 0,1723 |

**Tabela 5.3 -** Resultados

# 6 Conclusões

De acordo com os resultados apresentados, é possível concluir que o uso de AGCs, para a otimização de parâmetros de RNAs, se mostrou-se eficiente. A união entre AGs e ACs conseguiu isolar os indivíduos de forma que as soluções encontradas nas pesquisas evolucionárias não fossem perdidas durante os ciclos reprodutivos, mantendo assim, a diversidade genética na população. Além do mais, as abordagens celulares estudadas mostraram suas vantagens e desvantagens na aplicação da busca evolucionária de RNAs. A combinação de AGCs e RNAs conseguiu obter redes com baixos índices de erro ao reduzir número de camadas e neurônios por camada. Quando comparados com o método NNGA-DCOD, o número médio de neurônios por arquitetura é um pouco maior, mas ainda assim é um numero considerado reduzido. Esse resultado já era esperado, uma vez que a função de aptidão do NNGA-DCOD é mais custosa e consegue favorecer a obtenção de redes com menor número de neurônios, enquanto a função de aptidão utilizada nos AGCs visa apenas reduzir o erro médio das redes. Após a coleta dos resultados, ficou claro que, mesmo sem utilizar uma função de aptidão, que favoreça a obtenção de redes compactas, os critérios adotados na configuração das PAFs e no processo de mutação ajudaram na obtenção de redes compactas.

| Resultados do Teste $f$ | | | | | | |
|---|---|---|---|---|---|---|
| Bases | algoritmos | | AGC-1 ASSINCRONO | | AGC-2 SÍNCRONO | |
| | | | BP | SCG | BP | SCG |
| C. Alemão | NNGA-DCOD | BP | > | - | > | - |
| | | SCG | - | > | - | > |
| Ionosfera | NNGA-DCOD | BP | > | - | < | - |
| | | SCG | - | > | - | > |
| Veículos | NNGA-DCOD | BP | < | - | < | - |
| | | SCG | - | > | - | > |
| Cavalos | NNGA-DCOD | BP | > | - | > | - |
| | | SCG | - | > | - | > |
| Diabetes | NNGA-DCOD | BP | < | - | > | - |
| | | SCG | - | > | - | > |
| Câncer | NNGA-DCOD | BP | > | - | > | - |
| | | SCG | - | > | - | > |
| Vidros | NNGA-DCOD | BP | < | - | > | - |
| | | SCG | - | > | - | > |
| Coração | NNGA-DCOD | BP | < | - | > | - |
| | | SCG | - | > | - | > |
| Sonar | NNGA-DCOD | BP | < | - | < | - |
| | | SCG | - | < | - | < |
| C. Australiano | NNGA-DCOD | BP | > | - | > | - |
| | | SCG | - | < | - | > |

**Tabela 6.1 -** Resultados do teste $f$

O uso de AGCs assíncronos mostrou, em alguns casos, capacidade de encontrar soluções que, provavelmente, só seriam encontradas após muitas gerações pelo AGC síncrono (como foi o caso das bases Ionosfera e Cavalos ao utilizar o algoritmo SCG). No entanto, a sua característica não síncrona acaba por dispensar possibilidades de cruzamento que apenas o AGC síncrono conseguiu obter. Dessa forma, apesar de ambos os AGCs conseguirem dar boa capacidade de manutenção a população, o AGC síncrono consegue fazer uma busca mais profunda dentro do espaço de soluções e, por isso, tem maiores chances de encontrar soluções adequadas aos problemas. Contudo, vale ressaltar que a aplicação de um AGC síncrono possui suas restrições, uma vez que ele requer mais tempo de busca para que boas soluções sejam encontradas, enquanto o AGC assíncrono é mais rápido e, caso encontre um bom nicho na população dentro de sua pesquisa, as possibilidades de obter um indivíduo com aptidão melhor aumentam, de forma a desenvolver um determinado local da população e aumentar as chances de encontrar soluções que só seriam encontradas após várias gerações. Assim, as duas técnicas apresentadas possuem suas características próprias, mas ambas conseguiram administrar com eficácia a evolução das populações graças à capacidade de dispersar lentamente as soluções através do AC, é possível aplicar uma maior manutenção da diversidade da população, evitando a convergência genética e proporciona equilíbrio entre a capacidade de exploração (exploration) e aproveitamento (exploitation).

De forma geral, os resultados dos AGCs, quando comparados com a busca manual, também conseguiram bons índices, com exceção das bases Veículo e Vidros, nas quais a busca manual conseguiu se aproximar um pouco dos resultados dos AGCs quando utilizaram o algoritmo BP e nas bases Câncer, Vidros e Sonar com o algoritmo SCG.

Os algoritmos de aprendizagem aplicados também merecem a devida atenção. De forma geral, o algoritmo BP é bastante sensível à quantidade de épocas de treinamento e, com isso, não obteve resultados muito significantes. Contudo, vale lembrar que foram utilizadas dez bases do repositório UCI e que oito destas são consideradas difíceis. Além disso, o número reduzido de buscas evolucionárias, em todas as populações, pode ter sido de grande impacto para que o resultado do algoritmo BP não fosse tão bom quanto esperado, uma vez que a quantidade reduzida de gerações também reduz as chances de encontrar os melhores parâmetros para as RNAs, que iriam suprir algumas necessidades do BP. Por outro lado, quando aplicado o algoritmo SCG, os resultados obtidos foram bons e conseguiram se aproximar dos resultados encontrados na literatura.

A busca automática por RNAs apresentou resultados significativamente melhores que a busca manual para a maioria das bases difíceis utilizadas, tais como a base Crédito Alemão, Veículos (tida como uma das bases mais difíceis), Cavalos e Crédito Australiano com o algoritmo SCG. De forma geral, o erro da busca manual com o algoritmo BP foi muito alto, porém condizente com uma busca manual feita em bases difíceis. Enquanto os erros obtidos com o algoritmo SCG foram, em alguns casos, mais próximos dos erros encontrados com a busca automática.

Um dos principais problemas encontrados pelos métodos apresentados é o seu elevado tempo de processamento. Assim como outras buscas evolucionárias, o tempo de processamento necessário para a aplicação dos AGCs em conjunto com as RNAs é demasiadamente grande. As principais causas para esse grande tempo é o uso da ferramenta Matlab, que facilita o desenvolvimento, mas consome muita memória e processamento; o conjunto de *loops* aninhados que são utilizados na pesquisa evolucionária e a forma de programação utilizada para os AGCs, uma vez que a implementação mais apropriada para esta técnica seria feita através de conceitos de programação paralela.

# Referências